\documentclass[letterpaper]{article} 
\usepackage{aaai2026}  
\usepackage{times}  
\usepackage{helvet}  
\usepackage{courier}  
\usepackage[hyphens]{url}  
\usepackage{graphicx} 
\usepackage{natbib}  
\usepackage{caption} 
\usepackage{algorithm}
\usepackage{algorithmic}
\usepackage{amsmath}

\usepackage{newfloat}
\usepackage{listings}
\DeclareCaptionStyle{ruled}{labelfont=normalfont,labelsep=colon,strut=off} 
\floatstyle{ruled}
\newfloat{listing}{tb}{lst}{}
\floatname{listing}{Listing}
\title{Positional Cognitive Specialization: Where Do LLMs Learn To Comprehend and Speak Your Language?}
\author{
    Luis Frentzen Salim\textsuperscript{\rm 1, 2},
    Lun-Wei Ku\textsuperscript{\rm 1},
    Hsing-Kuo Kenneth Pao\textsuperscript{\rm 2}
}
\affiliations{
    \textsuperscript{\rm 1}Institute of Information Science, Academia Sinica\\
    \textsuperscript{\rm 2}Department of Computer Science and Information Engineering, National Taiwan University of Science and Technology\\
    
    m11315804@mail.ntust.edu.tw, lwku@iis.sinica.edu.tw, pao@mail.ntust.edu.tw
}

\begin{document}

\maketitle

\begin{abstract}
Adapting large language models (LLMs) to new languages is an expensive and
opaque process. Understanding how language models acquire new languages and
multilingual abilities is key to achieve efficient adaptation. Prior work
on multilingual interpretability research focuses primarily on how trained
models process multilingual instructions, leaving unexplored the mechanisms
through which they acquire new languages during training. We
investigate these training dynamics on decoder-only transformers through the
lens of two functional cognitive specializations: language perception (input
comprehension) and production (output generation). Through experiments on
low-resource languages, we demonstrate how perceptual and productive
specialization emerges in different regions of a language model by running 
layer ablation sweeps from the model's input and output directions. Based
on the observed specialization patterns, we propose \textbf{CogSym}, a layer-wise 
heuristic that enables effective adaptation by exclusively finetuning a few early 
and late layers. We show that tuning only the 25\% outermost layers achieves
downstream task performance within 2--3\% deviation from the full finetuning
baseline. 
Unlike similar layer-selection methods, the proposed method requires no extra data 
or computation while retaining comparable performance, which is especially 
beneficial for low-resource languages.
\textbf{CogSym} yields consistent performance with adapter
methods such as LoRA, showcasing generalization beyond full finetuning. These
findings provide insights to better understand how LLMs learn new languages and
push toward accessible and inclusive language modeling.
\end{abstract}

\begin{links}
    \link{Code and extended version}{https://github.com/luisfrentzen/cognitive-specialization}
\end{links}

\begin{figure*}[t]
\centering
\includegraphics[width=0.975\textwidth]{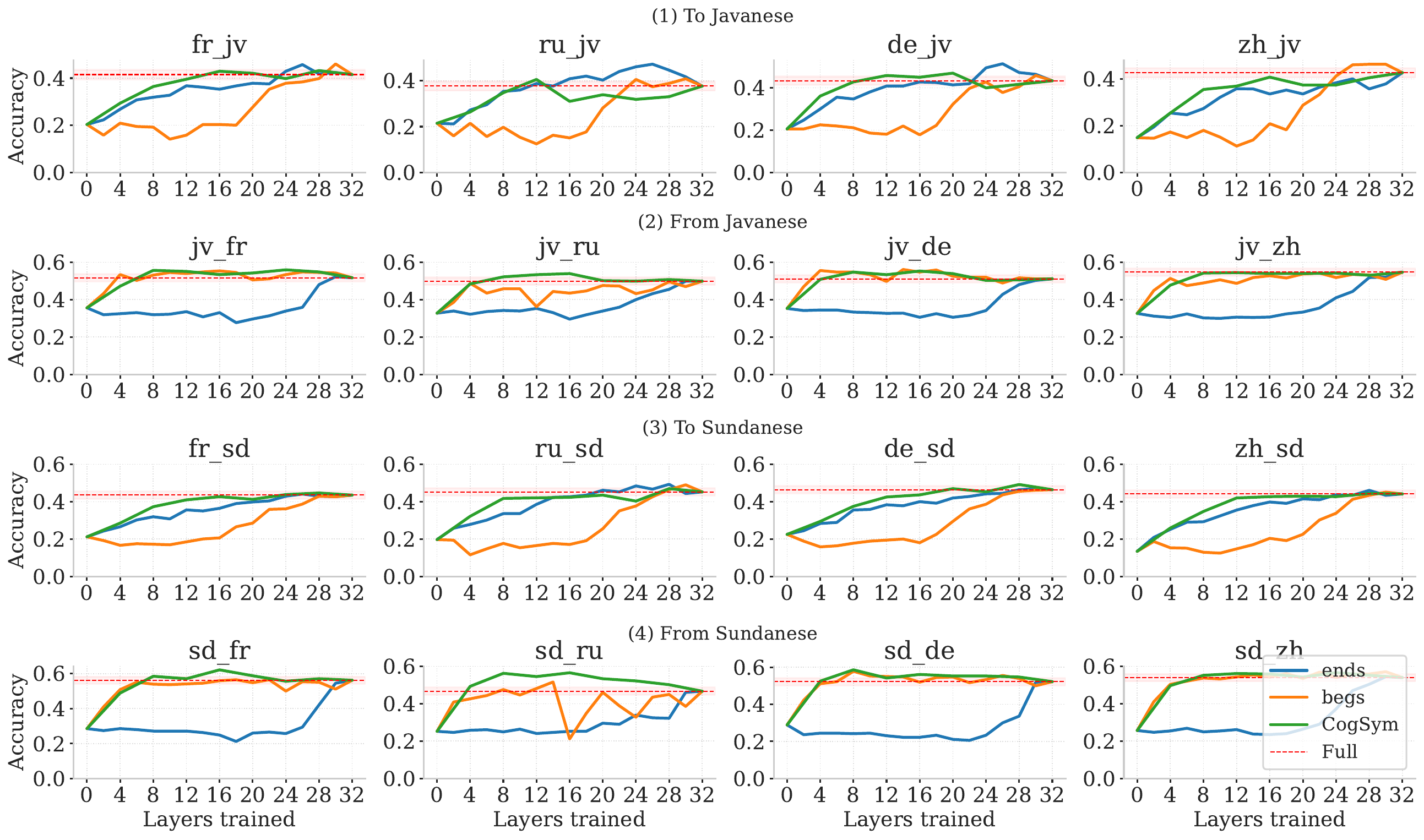} 
\caption{Word-level translation task performance of each ablation sweep, namely
front sweep (begs), rear sweep (ends), and both-ends or Cognitive-Symmetric (CogSym) sweep. Plots visualize
emerging patterns as each region is expanded.}
\label{mainlang}
\end{figure*}

\begin{figure*}[t]
\centering
\includegraphics[width=0.975\textwidth]{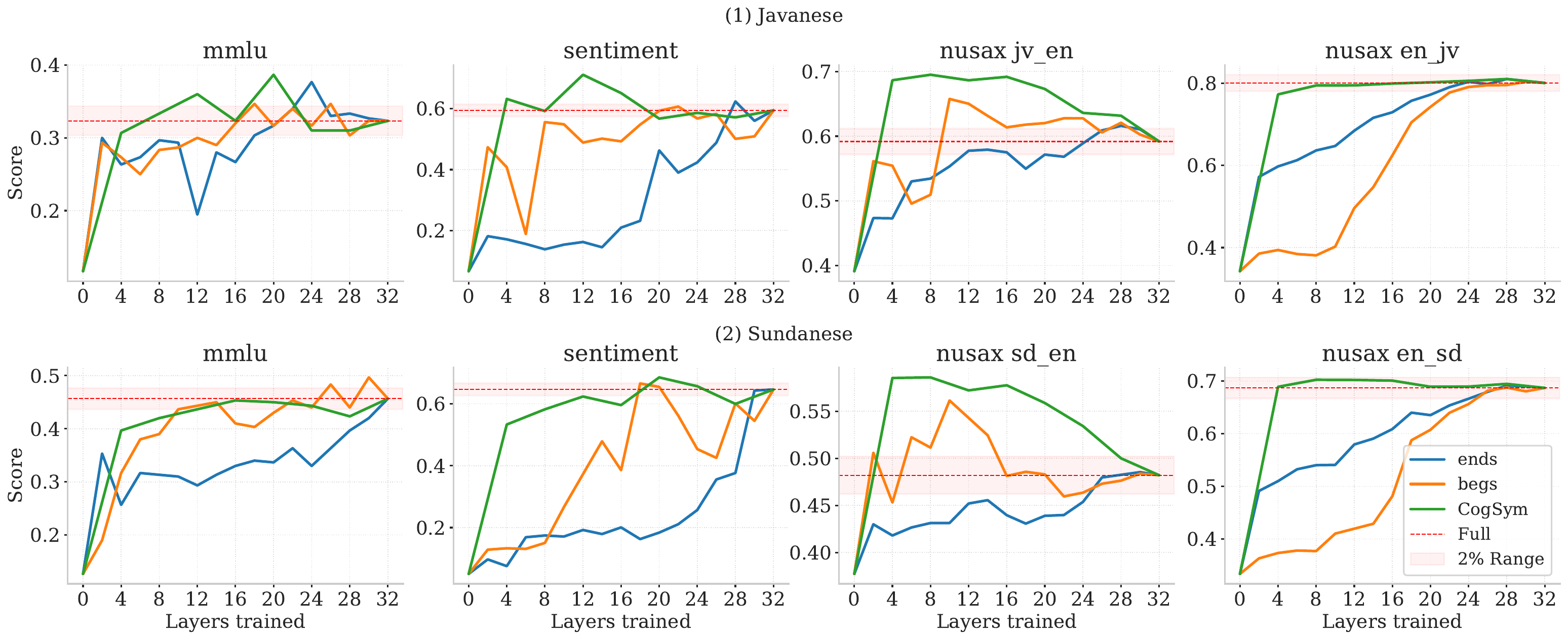} 
\caption{Downstream task performance of each ablation sweep}
\label{maintask}
\end{figure*}

\section{Introduction}

Large language models (LLMs) are central to many modern natural
language processing (NLP) tasks, demonstrating remarkable generalization
capabilities across diverse tasks and languages~\cite{NEURIPS2020_1457c0d6,
xue-etal-2021-mt5,
shi2022languagemodelsmultilingualchainofthought}. However, most are
English-centric, trained on massive English 
text~\cite{touvron2023llama2openfoundation, zhang2022optopenpretrainedtransformer}. This
focus limits the accessibility and benefits of the technology, especially for
speakers of low-resource languages. Despite efforts to bridge this gap,
proposed methods are often expensive. A key research challenge is thus to
understand the underlying architectural mechanisms of multilingualism to
develop more efficient and effective adaptation methods.

Prior work on multilingual interpretability has revealed an emerging
framework where multilingual processing occurs in three distinct stages: an
understanding phase in early layers to process the input language, an
intermediate thinking phase in the middle layers that operates on a shared
representation (either language-agnostic or closer to the model's
dominant language), and a generation phase in later layers for the output
language~\cite{Wendler2024-ar,Zhao2024-bg,Tang2024-fd,Schut2025-ef}. This
mirrors how polyglots manage multiple languages, where their internal
thoughts can operate in a different language than that used for immediate
speaking and hearing. These emergent specializations resemble 
language centers in the human brain, such as Broca's area for speech
production~\cite{broca} and Wernicke's area for language 
comprehension~\cite{wernicke1874aphasische}.

However, since prior research has predominantly focused on observing this
multilingual property on trained models, its behavior during the training
process and impact on effective adaptation remains underexplored. Understanding
its dynamics during training could reveal more efficient adaptation strategies
that target only those layers that develop the necessary functionality. Building
on these observations, we hypothesize that behind this behavior exist
specialized regions that are each responsible for a different cognitive
functionality to process multiple languages, akin to the different language
centers in the human brain. Intuitively, we can target only those regions that
yield the best return for our adaptation goals.

Current language adaptation methods for LLMs typically involve tuning 
  all of the internal layers.   
However, taking reference from human second language acquisition (SLA),
the acquisition of a new language does not necessitate a complete overhaul of
the entire cognitive system: indeed, it often involves acquiring only new linguistic
interfaces while preserving existing conceptual 
knowledge~\cite{Slobin1996-SLOFTA, thinkingemanuel}. Accordingly, we posit that efficient
language adaptation requires training only these interfaces in the front and
rear layers while preserving the weights that process shared representations in
the middle layers. To test this, we ablate consecutive layer ranges starting
from the front and rear ends of the model, sweeping in two-layer intervals.
We then probe the model's comprehension and generation abilities by evaluating
the trained models on translation tasks between the newly trained language and
languages known by the base model. Specifically, we examine tasks that require
understanding or comprehension of a target language, and tasks that require
generation or production of the target language, e.g., translation from known
languages to the target language (generation) and from the target language
to known languages (comprehension). We refer to the newly trained or target
languages as \textit{unknown} languages and existing languages as 
\textit{known} languages.

Our primary contributions in this work are threefold:

\begin{enumerate}
	 \item We observe the existence of what we term \textit{perceptual-productive specialization} 
  during the language adaptation process; this is responsible for the different phases of 
  multilingual processing, and is a phenomenon that reflects the language centers of the human
	 brain. 

	 \item We propose a simple yet effective positional heuristic, \textbf{CogSym} that is
	 training-method agnostic, proving effective for both full finetuning and
	 parameter efficient finetuning (PEFT) methods like low rank adaptation
	 (LoRA). 
     Moreover, it works without requiring any additional data, which is valuable for low-resource languages.
     We demonstrate that this heuristic achieves performance comparable
	 to full-model tuning by training as few as the 25\% outermost layers in
	 low-resource scenarios.
	 \item We characterize the key properties of this heuristic, providing novel
	 insight into the effective and resource-efficient adaptation of LLMs for
	 diverse linguistic contexts, especially for low-resource languages.
\end{enumerate}

\section{Related Work}

\textbf{Multilingual Language Models}. Modern language models have shown
excellent multilingual proficiency. Multilingual models such as 
mBERT~\cite{devlin2019bertpretrainingdeepbidirectional}, 
XLM-R~\cite{conneau2020unsupervisedcrosslingualrepresentationlearning}, 
mGPT~\cite{lin-etal-2022-shot}, 
PolyLM~\cite{wei2023polylmopensourcepolyglot}, and 
Aya~\cite{ustun2024ayamodelinstructionfinetuned} demonstrate how a single model
can effectively represent and process dozens of languages simultaneously.
However, they often exhibit a significant performance gap on lower-resource
languages due to their skewed pretraining corpora. Training a powerful model
from scratch for a low-resource language is also infeasible due to the scarcity
of high-quality, large-scale data. To address this, research has focused on
various strategies for language adaptation. These approaches typically involve
resource-intensive continued pretraining on target-language data or
finetuning on specific downstream tasks to transfer the model's existing
capabilities to a new linguistic context.

\textbf{Multilingual Interpretability}. A central challenge in the study of
LLMs are their black-box nature. The field of
interpretability aims to demystify the internal
mechanisms of these models to understand how they achieve complex capabilities.
Interpretability is crucial to developing robust, reliable, and efficient
models. Multilingual models are no exception to this.

Recent interpretability work on multilingual LLMs has uncovered a consistent
narrative suggesting that multilingual LLMs often follow a three-stage process
that maps onto their architecture. For instance, \cite{Zhao2024-bg} shows how
early layers and later layers process mostly non-English tokens whereas 
middle layers operate mostly on English tokens. Given this observation, they
propose a workflow where models first understand an input in its source
language, then solve the task using an internal representation heavily
influenced by high-resource languages, before generating an output in
the target language. \cite{Tang2024-fd} explore the idea of language-specific
neurons, finding that those neurons are responsible for processing the
corresponding language and are disproportionately concentrated in the front and
rear layers of the model. \cite{Wendler2024-ar} explore the layers latent
representations using the logit 
lens~\cite{nostalgebraist2020interpreting}, demonstrating that models like Llama
often pivot through a shared representation in their middle layers before
generating in the target language.

\textbf{Layer-Selective Finetuning}. A common strategy for efficient
adaptation is layer-selective tuning, which trains only a select subset of a
model's most important layers to reduce computational costs while maintaining
performance. \cite{lee2023surgical} introduces surgical finetuning, which
selectively trains layers to adapt to different types of distribution shifts
during transfer learning on computer vision models. In a similar study on
encoder-only language models, 
\cite{Lodha2023-bs, wang2025floefisherbasedlayerselection} leverages Fisher
information matrix (FIM) computed on a small sample of data to rank layers
based on their significances. 
\cite{zhang-etal-2024-investigating} experiments
with the Shapley value to measure LLM layer importance, discovering that their
performance relies on a few ``cornerstone'' layers that severely damage
performance when removed. 
\cite{qing-etal-2024-alphalora} introduces AlphaLoRA,
a method leveraging the Heavy-Tailed Self-Regularization theory to select
the best layer to train based on the model weights.

\section{Methodology}

This study presents two low-resource languages: Javanese and Sundanese. To
assess the effectiveness of our training, for some tasks we include
English, German, Chinese, Russian, and
French, all higher-resource languages that are well-represented in the
pretraining data of the base models and act as a measure to evaluate
against in tasks such as machine translation.

\subsection{Datasets}

\textbf{Training Data}. To facilitate language adaptation, we used a version of
the Alpaca instruction dataset presented by~\cite{alpaca, upadhayay2024taco},
which was translated using Google Translate into multiple low-resource
languages. This dataset contains 52,000 rows of diverse instructions. The
dataset's nature and size provide us with a wide range of vocabulary and text
semantics in the form of instructions. Its modest size also allows us to
simulate a low-resource adaptation scenario.

\textbf{Evaluation Data}. To evaluate our adaptation results, we defined a
comprehensive suite consisting of four different tasks designed to measure
training performance across different granularities. Each dataset includes
evaluation data for both target languages.

\begin{itemize}
	 \item \textbf{Word-level Translation}. We assess vocabulary acquisition
	 using a word-level translation task. The dataset used is adapted from the
	 data presented by~\cite{Wendler2024-ar}. The dataset consists of words that
	 map to single tokens in the Llama 2 tokenizer. The Javanese and Sundanese
	 subsets---not included in the original dataset---were translated
	 using Google Translate. In a 4-shot manner, we evaluate the model's ability to translate
	 from the trained target language to the languages present in this
	 dataset in both directions. Performance is measured by accuracy. 
    
	 \item \textbf{Machine Translation}. This task enables us to observe
	 the model's ability to handle full grammatical sentences. We employ the
	 NusaX dataset~\cite{winata-etal-2023-nusax}, which provides parallel
	 corpora between English and both target languages. We evaluate the
	 model's machine translation ability in both directions, from English to the
	 target language and vice versa. The prompt utilizes a 2-shot setting. We
	 evaluate performance using COMET-22~\cite{rei-etal-2022-comet} as it is
	 consistently present as an official metric in the latest and 
  past WMT Shared Task~\cite{kocmi-etal-2024-findings}. 
    
	 \item \textbf{Sentiment Analysis}. We evaluate semantic understanding using
	 the sentiment analysis component of the NusaX dataset, which contains texts
	 annotated with three sentiment labels (positive, neutral, negative),
	 allowing us to assess the model's ability to capture nuanced meaning in the
	 target languages. We evaluate in a 3-shot manner, which includes one shot
	 for each class. We report the macro F1-score. 

	 \item \textbf{IndoMMLU}. We use IndoMMLU~\cite{koto-etal-2023-indommlu} to
	 assess general language comprehension. This benchmark contains
	 multiple-choice questions with up to five options from Indonesian educational
	 curricula spanning primary school through university entrance exams. We
	 specifically use the primary school Javanese and Sundanese language
	 subjects, which focus on basic language comprehension, grammar, and
	 vocabulary. Higher grades tend to include culture-specific knowledge that
	 would not be acquired through our instruction-based training. We evaluate
	 in a 2-shot manner. We measure performance using accuracy. 
\end{itemize}

We report the average score over multiple evaluation runs for all tasks. 

\subsection{Experimental Setup}

To investigate how the specialized cognitive regions are distributed across the
model layers, we conduct a systematic ablation sweep where we train a model
while unfreezing only consecutive series of layers, each acting as a ``region''
given a fixed trainable layer budget $k$ with $k \in \{2, 4, 6, \ldots, L\}$ where
$L$ is the total number of layers in a decoder-only transformer model. 
We train the LM head and the embedding layer as part of the sweep, however, 
they are not counted towards $k$.
Our primary experiments allocate $k$ in three ways: 

\begin{itemize}
\item \textbf{Front Layer Sweep}: We start with $k=2$ and unfreeze only $k$
layers nearest to the input (front of the model). Then, we progressively expand by
an interval of $2$ until we reach the maximum value of $k$, totaling $16$
training variations for the 32-layered model. We also train the embedding layer
as part of the front sweep.

\item \textbf{Rear Layer Sweep}: We start with $k=2$ and unfreeze only $k$
layers nearest to the output (rear of the model). Likewise, we expand by
an interval of $2$ until we reach the maximum value of $k$, totaling $16$
training variations. We also train the LM head as part of the rear sweep.

\item \textbf{Both/Cognitive-Symmetric (CogSym) Sweep}: We start with $k=4$ and split $k$
equally between the front and rear regions, training the first $k/2$ and last
$k/2$ layers. We expand by $4$ until the two
regions meet in the middle and we reach the maximum value of $k$, totaling $8$
training variations. In this sweep, we train both the LM head and the embedding layer.
\end{itemize}

This design allows us to study how the position and size of the regions affect
model performance on the evaluation set. We also compare to three other ways to
allocate $k$, by leveraging:
\begin{itemize}
\item \textbf{Language-Specific Neuron (LSN) Count}: We adapt the method 
proposed by \cite{Tang2024-fd}, which select neurons by ranking its activations 
over a monolingual corpus.
Since LSNs are known to accumulate in certain layers, we rank layers by 
the number of LSNs they possess. We use the detected neurons of the Llama 2 7B 
model for the Chinese, French, Indonesian, and Spanish languages provided 
by the paper authors. 

\item \textbf{Layer-Wise Fisher Information Matrix (FIM)}: We employ the layer
selection method proposed by \cite{Lodha2023-bs}, which uses FIM to identify the most
informative layers. This method quantifies the impact of parameter changes on a
model's prediction by computing the diagonal elements of the FIM and ranking
layers accordingly. We directly use the FIM score calculation as defined in
their paper: 
\begin{align}
F_\theta &= E_{x \sim p(x)}\left[E_{y \sim p_\theta(y|x)}\left[\nabla_\theta \log p_\theta(y|x) \right.\right. \nonumber \\
&\quad \left.\left. \nabla_\theta \log p_\theta(y|x)^T\right]\right].
\end{align}
To compute our layer-wise FIM, we sample 250 rows from our finetuning dataset.

\item \textbf{Heavy-Tailed Self-Regularization (HT-SR)}: 
We use the $PL\_Alpha\_Hill$ metric proposed by \cite{qing-etal-2024-alphalora}
to measure layer importance. It is defined by taking the eigenvalues of a 
layer's weight correlation matrix, fitting a power-law distribution to the 
heavy-tailed part of the eigenvalue spectrum, and estimating the power-law 
exponent using the Hill estimator. Layers with higher values correspond to 
less heavy-tailed distributions which require more tuning.

\end{itemize}

\subsection{Training Details}
We conducted all experiments using RTX 3090/4090 24GB and A6000/A6000 Ada 48GB
GPUs. We used the Unsloth library and conducted all training in bfloat16
precision. All training was optimized with AdamW 8-bit with a batch size of
$16$, trained for one epoch with a fixed random seed of $42$ for
reproducibility. All LoRA runs used $r=128$.

\begin{figure}[t]
\centering
\includegraphics[width=0.95\columnwidth]{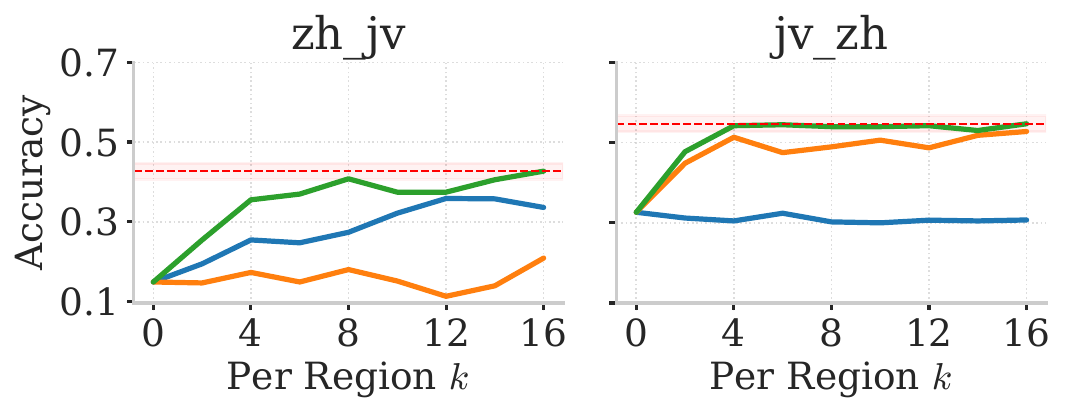} 
\caption{Comparison between each sweep strategy with regard to singular region budget $k$}
\label{comparison}
\end{figure}

\begin{figure}[t]
\centering
\includegraphics[width=0.65\columnwidth]{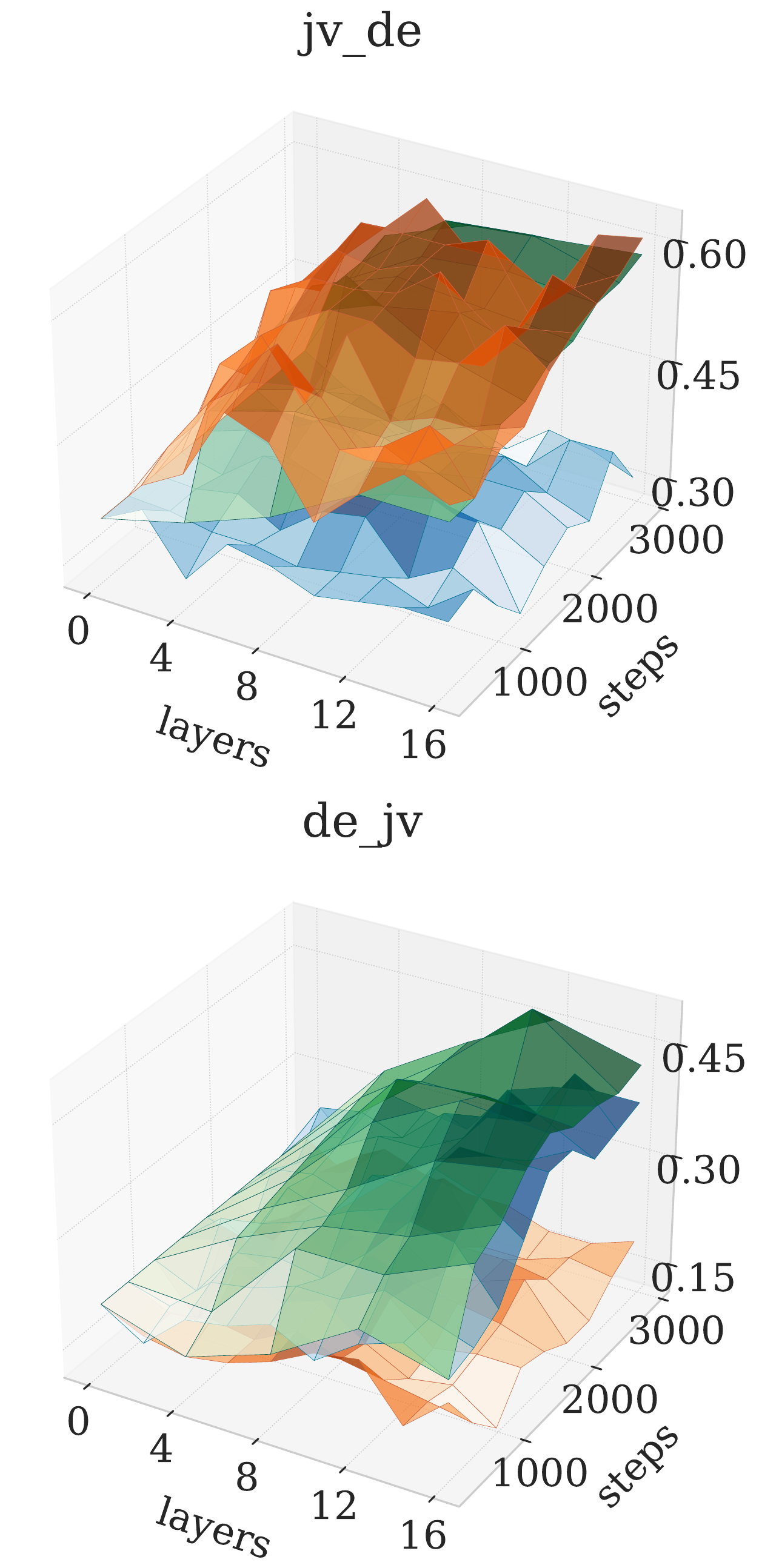} 
\caption{3D view of word-translation task plot with training steps as an
additional axis for German--Javanese pair}
\label{3d}
\end{figure}

\section{Results and Discussion}

In this section we discuss the ablation training results and
analyze the emerging patterns. Different selection methods are distinguished
by a consistent color across all visualizations. The baseline performance of
the fully trained model is denoted by the dotted red line, and the base
(untrained) model performance by the dotted green line. We will focus our 
discussion on the Llama 2 7B model.

\subsection{Separation of Perceptual-Productive Specialization}

We plot the performance of the models trained in Javanese and Sundanese in
Fig.~\ref{mainlang} and~\ref{maintask}. We observe a clear progression in
task performance as the size of each region is increased. On translation tasks
from known languages to unknown languages (Fig.~\ref{mainlang}, first row),
performance is overwhelmingly dependent on the rear layers. Training only the
rear (ends) layers achieves high accuracy with just 10 layers, whereas training
the front (begs) layers yields no performance gain for the same budget $k$. 
Conversely, on translation from the unknown languages, the initial
layers are critical. Front-layer training rapidly approaches the fully trained
model's performance (red dotted line), whereas rear-layer training shows almost
no improvement until it reaches the front region. 

This pattern holds for 
other downstream tasks. Rear-layer training underperforms on IndoMMLU and
sentiment analysis, which typically require understanding of the case at
hand rather than complex generation. These results show a clear division
between perceptual and productive specialization in the front and rear layers,
respectively. Figure~\ref{3d} reveals how specialization develops across multiple checkpoints
during training. The performance gap emerges early in training and continues to
widen rapidly through the final checkpoint. This suggests these specializations
exist fundamentally, as they form quickly and intuitively within the first few
hundred steps. 

Although we observe isolated specialization in those regions, we also find that
when trained together, the performance of both regions is complementary.
As shown in Fig.~\ref{comparison}, we find that when plotted
with respect to the singular region budget $k$, training both regions often
outperforms training a region alone. Our proposed method, Cognitive-Symmetric Tuning (\textbf{CogSym}) leverages these observations by training both regions 
simultaneously with symmetric allocation of $k$. With \textbf{CogSym} we achieve performance
matching the fully trained model using only a small subset of the total 32
layers. This approach consistently matches or outperforms the baseline with
only 25\% of the layers trained. This result strongly supports our hypothesis
that efficient language adaptation can be achieved by training only the
linguistic interfaces. We also find that this targeted adaptation does not negatively 
impact the model's capabilities on other languages, which is one of the main 
concerns during language adaptation.

Since we observe consistent patterns in both Javanese and Sundanese, we focus
our subsequent analysis on Javanese.  

\subsection{Training Layer Position Matters}

\begin{figure}[t]
\centering
\includegraphics[width=0.95\columnwidth]{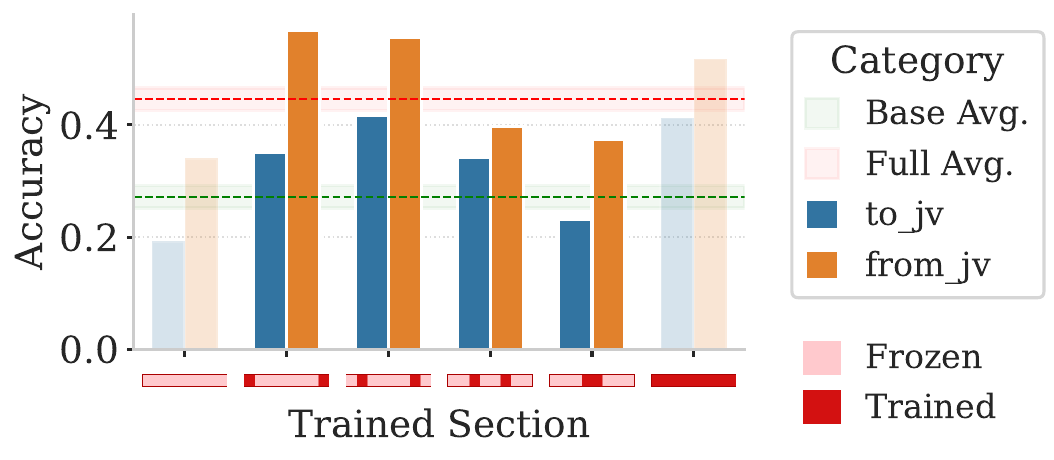} 
\caption{Word translation task performance of 4-position variant with $k=8$}
\label{closing}
\end{figure}

We now investigate whether this specialization requires distributing $k$ at 
the model's extreme ends. We fix a layer budget of $k=8$ (which
performed well in our initial experiments) and test four different position
configurations, each allocating four layers to the front and four to the rear regions
of the model. Figure~\ref{closing} illustrates our positional ablation study.
The leftmost configuration trains the eight outermost layers (0--3 and 28--31), whereas
subsequent configurations shift these regions inward by 4-layer increments,
eventually meeting at the center (layers 12--19). The red bars below each
configuration visualize the trained layer positions.

We observe that performance degrades as trained layers move away from
the model's extremities. This is most pronounced on the word translation task illustrated in
Fig.~\ref{closing}, the outermost configuration achieves
near-baseline performance across all source languages, whereas the centermost
configuration scores similarly to an untrained model. This holds 
across all evaluated tasks, though the magnitude of the performance
difference varies. This further supports the perceptual/productive regions discussed in the previous section.

\subsection{Comparison with Other Layer Selection Methods}

\begin{figure}[t]
\centering
\includegraphics[width=0.95\columnwidth]{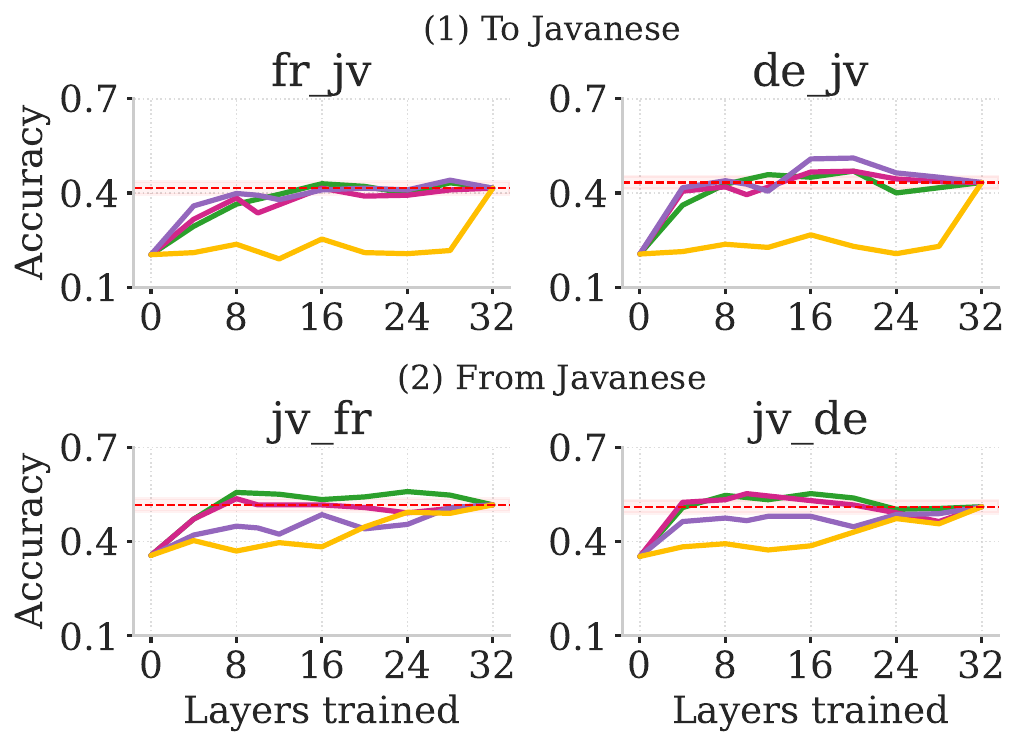} 
\caption{Word-level translation task performance comparison between
CogSym (green), FIM (purple), LSN count (magenta), and AlphaLoRA (yellow)}
\label{fishlsnlang}
\end{figure}

\begin{figure}[t]
\centering
\includegraphics[width=0.95\columnwidth]{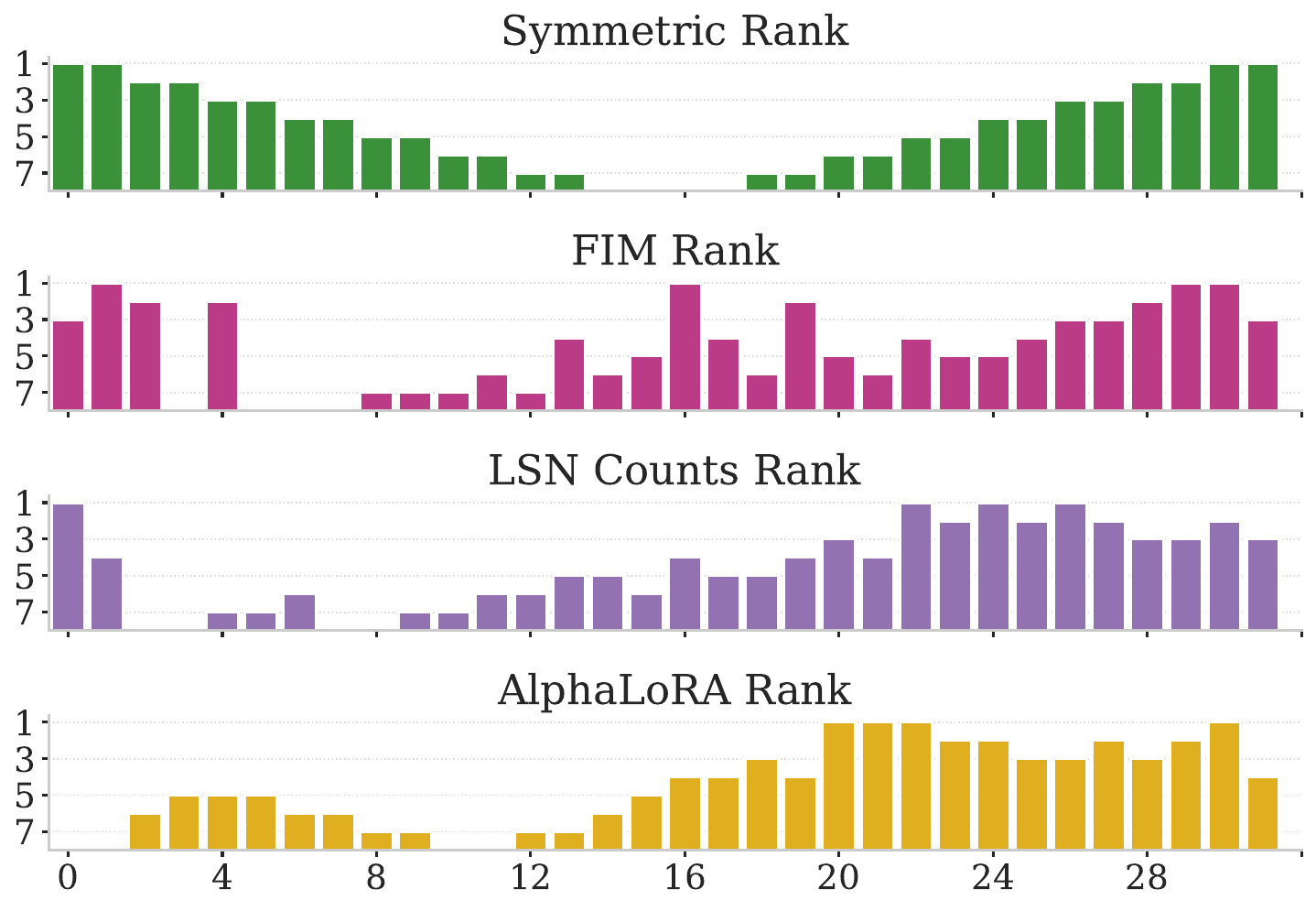} 
\caption{Full distribution of each layer budget allocation method,
higher bars indicate greater importance}
\label{ranks}
\end{figure}

We further evaluate \textbf{CogSym} with other layer selection methods, more
specifically by utilizing FIM, LSN counts, and AlphaLoRA to rank layers. 
Figure~\ref{fishlsnlang} show that \textbf{CogSym}
(green) and FIM-based selection (magenta) achieve comparable performance. In contrast,
LSN-based method and AlphaLoRA performs poorly.
This disparity stems from their skewness towards the
rear region, whereas \textbf{CogSym} and FIM prioritize both front and rear regions
more equally. The complete rank distribution of each method is presented in
Fig.~\ref{ranks}.

\textbf{CogSym} achieves results that are highly competitive
with the far more complex FIM method. The convergence of these two
philosophically distinct approaches provides strong evidence for the
fundamental importance of the model's extremities. 
Notably, unlike the other methods, our approach does not rely on additional 
data or computation, which is a valuable trait for adapting low-resource languages.

\subsection{Training Method Agnosticism}

\begin{figure*}[t]
\centering
\includegraphics[width=0.975\textwidth]{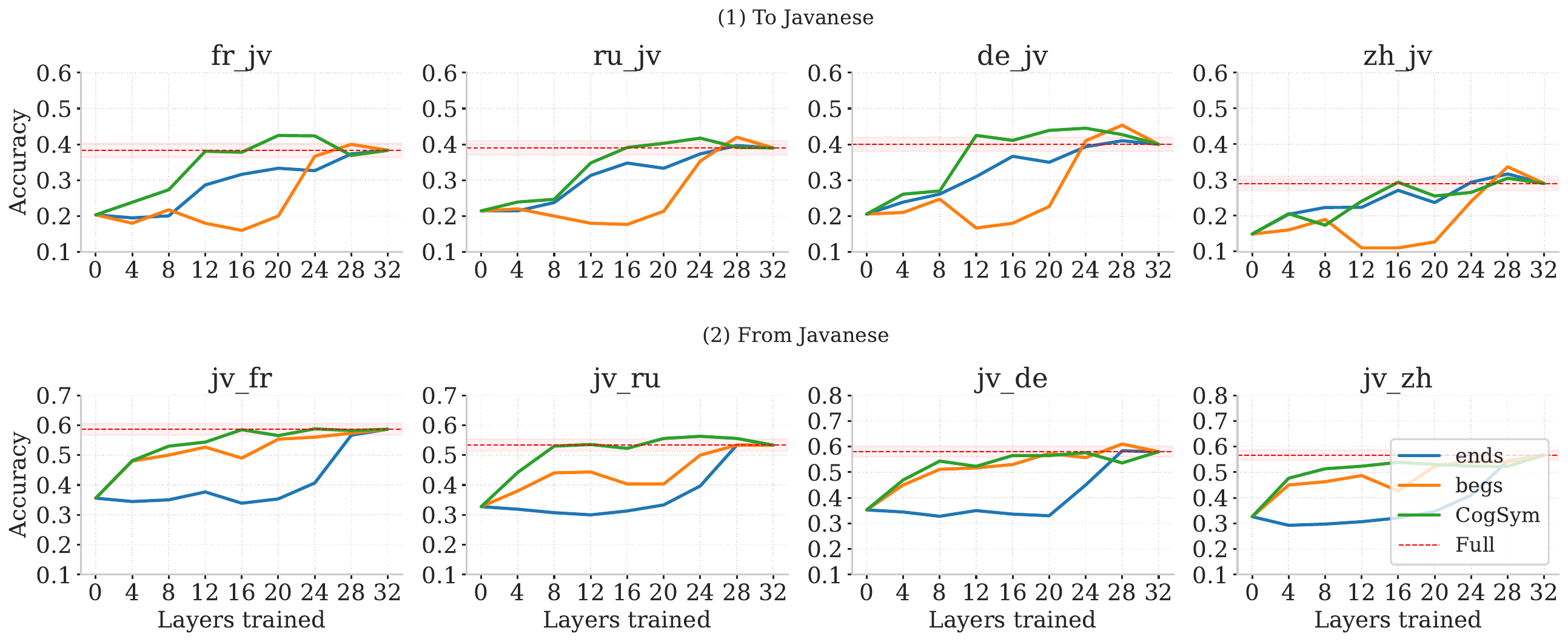} 
\caption{Word-level translation task performance of LoRA, which 
shows consistent patterns with full finetuning}
\label{loralang}
\end{figure*}

To investigate whether our functional specialization findings are inherent to the
model architecture rather than the training method, we replicated our
experiments with LoRA. Figure~\ref{loralang} shows \textbf{CogSym}
performance on LoRA. They demonstrate patterns that are remarkably consistent 
with those of full finetuning. The same functional specialization emerges: front layers
excel at comprehension-focused tasks, rear layers excel at generation-focused tasks,
and the symmetric approach effectively captures both capabilities. The
performance trajectories follow a nearly identical pattern. This experimental
result is evidence of training-method agnosticism: the critical factor is not how
weights are updated but where in the architecture the updates occur. This
suggests that the perceptive and productive interfaces represent fundamental
principles of the multilingual models. 

\subsection{Sequential Training}

\begin{figure}[t]
\centering
\includegraphics[width=0.95\columnwidth]{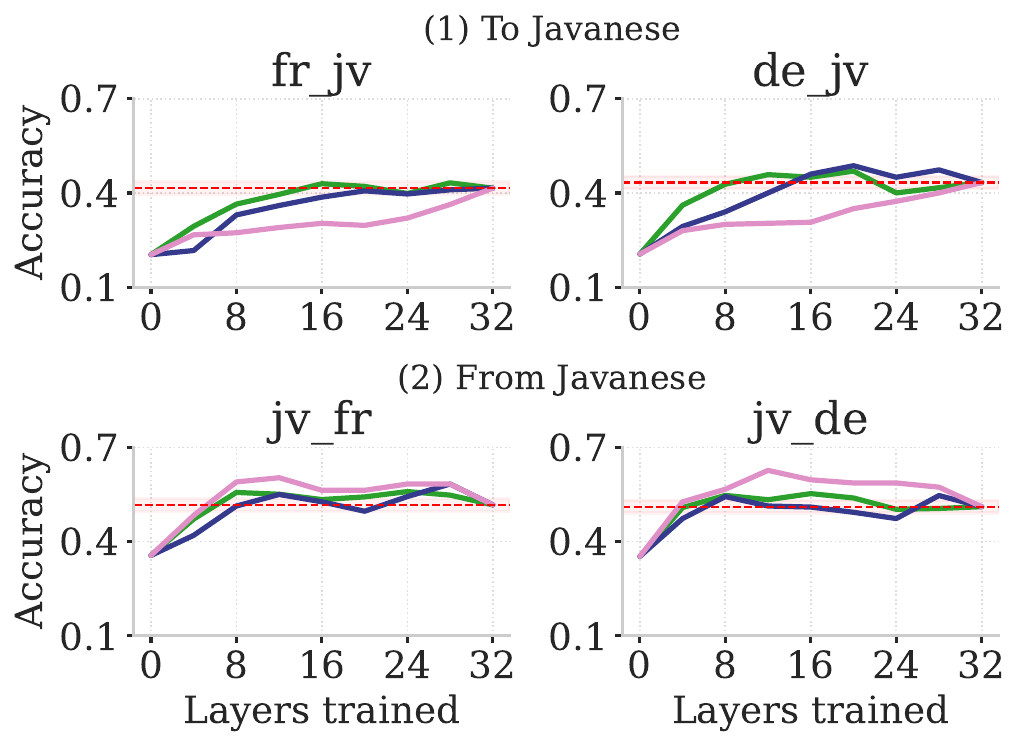} 
\caption{Word-level translation task performance comparison between
CogSym (green), front-first training (dark purple), and
rear-first training (pink)}
\label{seqslang}
\end{figure}

We ask whether the two regions can be trained separately rather than simultaneously in a single training run.
Since we have two separate regions,
training sequentially cuts memory usage by 50\%, which helps scenarios
where compute memory is limited. We tested two orderings: training the front
layers first, then the rear (beg\_end), and training the rear layers first,
then the front (end\_beg). Figure~\ref{seqslang} reveals that
training sequentially in this case works with minimal performance
degradation. But we also observe that recently trained regions
exhibit less performance degradation than regions trained
earlier. Front-first training yields minimal performance loss on
generation-focused tasks such as translation towards the unknown language, whereas
rear-first training suffers an almost 10\% degradation in performance.
Conversely, for comprehension-focused tasks, rear-first ordering performs
optimally. Perhaps catastrophic forgetting is a factor
during sequential training; we leave a more thorough investigation
of such dynamics for future work.

\section{Conclusion}

We seek to move beyond black-box approaches to language adaptation
and provide a clearer model of the underlying architectural dynamics connecting
it to human cognitive concepts. We demonstrate that the front and rear ends of
a multilingual language model are specialized to govern distinct cognitive
functionalities, which we term \textit{perceptual} and \textit{productive} specialized regions,
respectively. Our experiments reveal how the perceptual region governs
language comprehension and the productive region handles language generation,
similar to the distinct cognitive subareas of the human
brain, mirroring their corresponding neurological roles. Our observation
demonstrates that training beyond ten layers in these regions often
yields no additional performance gains on tasks requiring their specific
specialization. However, although seemingly isolated from each other, each region
is observed to complement the other when trained together in a way that boosts
performance in both specializations. Our proposed method \textbf{CogSym} leverages
this concept, enabling effective adaptation by training only the outermost layers.

We link \textbf{CogSym}'s performance to human SLA.
When humans learn new languages, they typically do not need to rewire their
core cognitive system; rather, they learn ``interfaces'' that enable
comprehension and communication in the new language. Previous study also found
that language-specific thinking patterns acquired through native/first
language learning during childhood is likely to prevent adult learners from attaining
target-like thinking patterns in the new language, which emphasizes acquiring
new linguistic packaging for thoughts instead of new thought patterns
themselves~\cite{thinkingemanuel}. These results demonstrate how we can rethink
language adaptation by mimicking how human cognitive functions leverage
existing linguistic foundations during language acquisition.

Practically, \textbf{CogSym} presents potential data and compute resource savings
of up to 75\% and beyond when combined with PEFT methods while achieving
minimal performance loss. This could help increase plasticity, by finding sub-networks
to help adapt languages at a lower cost and with fewer data points, since
training the smaller network often means we can use fewer data. This approach
offers significant advantages for low-resource language adaptation, enabling
cost-effective training with limited data by focusing on small, specialized
regions. The impact is particularly profound for low-resource language
communities, where data scarcity and computational constraints often coexist,
pushing for more accessible and inclusive language modeling. 

\section{Future Directions}
Whereas our study explores two of the most fundamental cognitive functionalities,
future studies could explore tasks that involve more complex
specializations such as reasoning or math abilities. For instance, the more
language-agnostic middle layers (a component not explored in this study) might hold 
more importance for those types of tasks.

\appendix
\section{Limitations}
Our experiment focuses on decoder-only transformers and two low-resource
languages from the Austronesian language family. Further work might be needed
to expand the analysis of this phenomenon towards other language models or
non-transformer architectures.

\bibliography{aaai2026}

\end{document}